# The Complexity of Approximately Solving Influence Diagrams

**Denis D. Mauá**
IDSIA
Switzerland, CH 6928
denis@idsia.ch

**Cassio P. de Campos**
IDSIA
Switzerland, CH 6928
cassio@idsia.ch

**Marco Zaffalon**
IDSIA
Switzerland, CH 6928
zaffalon@idsia.ch

## Abstract

Influence diagrams allow for intuitive and yet precise description of complex situations involving decision making under uncertainty. Unfortunately, most of the problems described by influence diagrams are hard to solve. In this paper we discuss the complexity of approximately solving influence diagrams. We do not assume no-forgetting or regularity, which makes the class of problems we address very broad. Remarkably, we show that when both the treewidth and the cardinality of the variables are bounded the problem admits a fully polynomial-time approximation scheme.

## 1 INTRODUCTION

Influence diagrams are well-known graphical models of decision making under uncertainty which allow for compact and intuitive representation of complex situations, and relatively fast inference [7–9, 17].

Most of the algorithms for inference with influence diagrams require a total ordering over the decisions and unlimited memory. The first assumption is called *regularity* and is graphically equivalent to the existence of a directed path comprising all decision variables in the diagram. The second assumption, called *no-forgetting*, states that decisions are made in light of all previously disclosed information. Graphically, it implies that the parent set of a decision node contains all previous decision nodes and their parents. These conditions allow an optimal solution to be obtained by dynamic programming [19], but impose an inherent exponential worst-case complexity, since the output might include a decision function over exponentially many values (e.g., a table prescribing a value for the last decision for each one of the exponentially many assignments of the other decision variables).

There are two common approaches to avoid the exponential complexity. The first one is to insert arcs entering decision variables in a way that makes them insensitive to the previous decisions, thus decreasing the size needed to represent its corresponding solution [8, 14]. This, however, implies observing quantities that were initially deemed unobservable, which might be undesirable or even unfeasible. The second approach is to relax no-forgetting and work with a limited memory, that is, to assume that not all previous decisions and observations are known and considered when making a decision [10, 13, 20]. Graphically, it corresponds to dropping arcs until the maximum size of a decision function becomes manageable. The latter approach, which gives rise to limited memory influence diagrams [10], is additionally justified by situations where decisions have to be taken independently of one another, as in team decision analysis [1, 6]; for such cases, some of the no-forgetting arcs are inherently absent.

Removing arcs from the original graph can make the problem harder. This essentially occurs because the problem might no longer be solvable by standard dynamic programming approaches. Zhang, Qi, and Poole [21] and more recently Lauritzen and Nilsson [10] determined sufficient conditions under which even influence diagrams that violate no-forgetting can be solved exactly by dynamic programming. Any diagram of bounded treewidth meeting these conditions can be thus solved efficiently. As de Campos and Ji [4] showed, however, even structurally very simple cases can fail to meet these conditions and be hard to solve. In fact, we have recently shown that even singly connected diagrams of treewidth equal to two, with decision variables having no parents and variables taking on a bounded number of states are NP-hard to solve, and that the problem is inapproximable if the cardinality of the variables is unbounded [12].

In this paper, we resume our investigation and analyze the impact of the number of value nodes and variable cardinality on the theoretical complexity of solving (limited memory) influence diagrams. We do not assume no-forgetting or regularity. We show that if the treewidth is bounded, the number of value variables do not affect the asymptotic complexity. Most importantly, we show that if the cardinalities of the variables are also bounded, the problem admits a fully polynomial-time approximation scheme.

## 2 INFLUENCE DIAGRAMS

To help introduce the notation and illustrate concepts, consider the following example of a decision problem where an agent first needs to select one among the actions $d_1^{(1)}, \ldots, d_1^{(i)}$, which causes one of $c_1^{(1)}, \ldots, c_1^{(j)}$ outcomes to obtain with probability $P(c_1|d_1)$, and generates a reward of $U(c_1)$. Then, without observing the value $c_1$, a second agent needs to select one among the actions $d_2^{(1)}, \ldots, d_2^{(k)}$, which generates one of the outcomes $c_2^{(1)}, \ldots, c_2^{(\ell)}$ with probability $P(c_2|c_1, d_2)$ and a reward of $U(c_2)$. The overall utility $U(d_1, c_1, d_2, c_2)$ of the agents' actions their corresponding outcomes is given by the sum of the intermediate rewards, that is, $U(d_1, c_1, d_2, c_2) = U(c_1) + U(c_2)$.

In the language of influence diagrams, the quantities and events of interest are represented by three distinct types of variables. *Chance variables* represent events on which the decision maker has no control, such as outcomes of tests or consequences of actions; *decision variables* represent the alternatives a decision maker has at each decision step; finally, *value variables* are used to represent immediate rewards. In the example, we can represent the decision of the first agent, its outcome, and the corresponding reward by a decision variable $D_1$, a chance variable $C_1$ and a value variable $V_1$, respectively. Similarly, we can represent the second decision, its outcome and the corresponding reward by a decision variable $D_2$, a chance variable $C_2$ and a value variable $V_2$. Regarding the notation, we denote variables by capital letters and identify them with the set of values they can assume; generic and particular values are denoted in lower case. For instance, a generic action of the first agent is represented as $d_1 \in D_1 = \{d_1^{(1)}, \ldots, d_1^{(i)}\}$, while the corresponding set of possible rewards (which depend on the outcome $c_1$) is given by $V_1 = \{U(c_1^{(1)}), \ldots, U(c_1^{(j)})\}$.

The functional dependencies between the variables in the model can be compactly represented by a directed acyclic graph (DAG) whose nodes are in a one-to-one correspondence to the (decision, chance and value) variables of the problem. For simplicity, we identify nodes with their associated variables. Hence, we can talk about the parents $\mathtt{Pa}(X)$ of a variable $X$, its children $\mathtt{Ch}(X)$, or yet its family $\mathtt{Fa}(X) = \mathtt{Pa}(X) \cup \{X\}$ of $X$. The arcs entering decision variables represent the set of variables whose values will be known by the time the corresponding decision is made, either because they are observed at this time (e.g., an outcome of a previous decision or an action taken by a different agent) or remembered (e.g., a previous decision or observation). The problem in the example can be depicted as a DAG in Figure 1 (as usual, chance, decision and value variables are represented by ovals, squares and diamonds, respectively). The existence of the dashed arc connecting $D_1$ to $D_2$ depends on whether the decision of the first agent is known by the second agent at the time the latter acts. It is present if the second agent knows (and takes into account) the decision of the first agent, and it is absent otherwise.

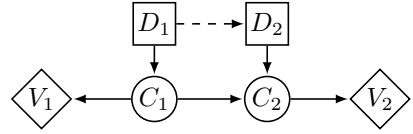

Figure 1: Influence diagram for the problem in the example.

An influence diagram is a tuple $(\mathcal{C}, \mathcal{D}, \mathcal{V}, \mathcal{G}, \mathcal{P}, \mathcal{U})$, where $\mathcal{C}$, $\mathcal{D}$ and $\mathcal{V}$ are the sets of chance, decision and value variables, respectively, $\mathcal{G}$ is a DAG over $\mathcal{C} \cup \mathcal{D} \cup \mathcal{V}$, $\mathcal{P}$ is a set containing one (conditional) probability distribution $P(C|\mathtt{Pa}(C))$ per chance variable $C$ in $\mathcal{C}$, and $\mathcal{U}$ is a set containing one reward function $U(\mathtt{Pa}(V))$ per value variable $V$. Thus, an influence diagram specifies both a *joint probability distribution* $P(\mathcal{C}|\mathcal{D}) = \prod_{C \in \mathcal{C}} P(C|\mathtt{Pa}(C))$ of the outcomes conditional on the decisions, and a *utility* $U(\mathcal{C}, \mathcal{D}) = \sum_{V \in \mathcal{V}} U(\mathtt{Pa}(V))$ over outcomes and decisions.

### 2.1 The Strategy Selection Problem

For any decision variable $D \in \mathcal{D}$, a *policy* is a conditional probability distribution $P(D|\mathtt{Pa}(D))$ prescribing a (possibly randomized) action for each state of the parents. We say that a policy $P(D|\mathtt{Pa}(D))$ is *pure* if for every assignment to the parents it assigns all the mass to a single action $d \in D$. A *strategy* is a set $\mathcal{S} = \{P(D|\mathtt{Pa}(D)) : D \in \mathcal{D}\}$ containing a policy for each decision variable. A strategy $\mathcal{S}$ induces a(n unconditional) joint probability distribution over decisions and outcomes by

$$P_{\mathcal{S}}(\mathcal{C}, \mathcal{D}) = P(\mathcal{C}|\mathcal{D}) \prod_{D \in \mathcal{D}} P(D|\mathtt{Pa}(D)),$$

and has an associated expected utility given by

$$\mathrm{E}_{\mathcal{S}} = \sum_{\mathcal{C}, \mathcal{D}} P_{\mathcal{S}}(\mathcal{C}, \mathcal{D}) U(\mathcal{C}, \mathcal{D}).$$

The primary use of an influence diagram is the *strategy selection problem*, which consists in finding a strategy $\mathcal{S}^*$ that maximizes the expected utility, that is, to find $\mathcal{S}^*$ such that $\mathrm{E}_{\mathcal{S}^*} \geq \mathrm{E}_{\mathcal{S}}$ for all $\mathcal{S}$. The value $\mathrm{E}_{\mathcal{S}^*}$ is called the *maximum expected utility* and is denoted shortly by MEU. Most algorithms for finding optimal strategies have complexity at least exponential in the treewidth of the influence diagram, which is a measure of its resemblance to a tree and is better formalized by the notion of tree decomposition.

A *tree decomposition* of an influence diagram is a tree-shaped graph where each node is associated to a subset of the chance and decision variables in the diagram. The decomposition satisfies the *family preserving* and the *running intersection* properties, namely, that the family of each decision and chance variable, as well as the parent set of each value variable, is contained in at least one set associated to a node of the tree, and that the graph obtained by dropping nodes that do not contain any given chance or decision variable is still a tree. The treewidth of a tree decomposition is the maximum number of variables associated to a node minus one. The treewidth of an influence diagram is the minimum treewidth of a tree decomposition of it.

For a fixed integer $k$, Bodlaender [2] showed that for any diagram one can in linear time either obtain a tree decomposition of treewidth at most $k$ or know that it does not exist. Hence, for any diagram of bounded treewidth we can obtain in linear time an optimal tree decomposition, that is, a tree decomposition whose treewidth equals the treewidth of the diagram. Moreover, any tree decomposition can be turned into a binary tree decomposition (i.e., one in which each node has at most three neighbors) of same treewidth in linear time [18]. Given a tree decomposition $\mathcal{T}$ with $m$ nodes, we denote by $\mathcal{X}_1, \ldots, \mathcal{X}_m$ the sets of variables associated to nodes $1, \ldots, m$, respectively. Thus, $\mathcal{X}_1 \cup \cdots \cup \mathcal{X}_m = \mathcal{C} \cup \mathcal{D}$.

Given an influence diagram, we can evaluate the expected utility of any strategy $\mathcal{S}$ in time and space at most exponential in its treewidth. Hence, if a diagram has bounded treewidth, we can obtain $\mathrm{E}_\mathcal{S}$ for any strategy $\mathcal{S}$ in polynomial time [9, Chapter 23]. Obtaining the optimal strategy in this way is however unfeasible for any moderately large problem, as the number of strategies is too high. In fact, de Campos and Ji [4] showed that even in diagrams of bounded treewidth the strategy selection problem is NP-hard. We have recently [12] strengthened their result by showing that the problem is already NP-hard in singly connected diagrams of treewidth equal to two and variables assuming at most three values. The result uses a reduction from the partition problem to a influence diagram whose underlying graph is a tree and contains a single value variable. One might then wonder whether allowing more than one value variable affects the difficulty of the problem. The answer, as the following result shows, is no.

**Theorem 1.** *For any influence diagram $\mathcal{I}$, there is an influence diagram $\mathcal{I}'$ containing a single value variable such that any strategy $\mathcal{S}$ for $\mathcal{I}$ is also a strategy for $\mathcal{I}'$ and obtains the same expected utility. Moreover, the treewidth of $\mathcal{I}'$ is at most the treewidth of $\mathcal{I}$ plus three.*

*Proof.* Let $\mathcal{I} = (\mathcal{C}, \mathcal{D}, \mathcal{V}, \mathcal{G}, \mathcal{P}, \mathcal{U})$ be an arbitrary influence diagram and consider, without loss of generality, a binary tree decomposition $\mathcal{T}$ of $\mathcal{I}$ such that for every value variable $V_i$ in $\mathcal{V}$ there is a leaf node in the tree whose associated set of variables is $\mathtt{Pa}(V_i)$.[1] Assume additionally that the tree is rooted at a node $r$ and that the leaf nodes $\ell_1, \ell_2, \ldots, \ell_q$ associated to the sets $\mathtt{Pa}(V_1), \ldots, \mathtt{Pa}(V_q)$, respectively, where $q = |\mathcal{V}|$ denotes the number of value variables, are ordered in such a way that they agree with an *in-order tree traversal* of the tree decomposition, that is, in a depth-first tree traversal of $\mathcal{T}$ rooted at $r$, the node $\ell_1$ precedes $\ell_2$, which precedes $\ell_3$, and so on. Let $\overline{U}$ and $\underline{U}$ be upper and lower bounds, respectively, on the reward functions in $\mathcal{U}$, with $\overline{U} > \underline{U}$.

Now consider a diagram $\mathcal{I}' = (\mathcal{C}', \mathcal{D}, \{V\}, \mathcal{G}', \mathcal{P}', \{U\})$, which contains a single value variable $V$ instead of the $q$ value variables of $\mathcal{I}$ and an augmented set of chance variables $\mathcal{C}' = \mathcal{C} \cup \{W_1, \ldots, W_q, O_1, \ldots, O_q\}$, where $W_1, \ldots, W_q, O_1, \ldots O_q$ are binary variables. Furthermore, the subgraph of $\mathcal{G}'$ obtained by considering only nodes in $\mathcal{C}$ and $\mathcal{D}$ is identical to $\mathcal{G}$ with the chance variables $W_1, \ldots, W_q$ replacing the value variables $V_1, \ldots, V_q$, respectively. Also, the variables $O_1, \ldots, O_q$ are arranged in a chain such that $W_1$ is the parent of $O_1$, $W_2$ and $O_1$ are the parents of $O_2$ and so forth, as in Figure 2(b). Each variable $W_i$, for $i = 1, \ldots, q$, is associated to a probability distribution $P(W_i|\mathtt{Pa}(V_i))$ such that $P(w_i^{(1)}|\mathtt{Pa}(V_i)) = (U(\mathtt{Pa}(V_i)) - \underline{U})/(\overline{U} - \underline{U})$. Each variable $O_i$, $i = 1, \ldots, q$, is associated to a probability distribution $P(O_i|O_{i-1}, W_i)$ (we assume $O_0 = \emptyset$) such that $P(o_i^{(1)}|o_{i-1}^{(1)}, w_i^{(1)}) = 1$ and $P(o_i^{(1)}|o_{i-1}^{(1)}, w_i^{(2)}) = (i-1)/i$ and $P(o_i^{(1)}|o_{i-1}^{(2)}, w_i^{(1)}) = 1/i$ and $P(o_i^{(1)}|o_{i-1}^{(2)}, w_i^{(2)}) = 0$ ($P(o_1^{(1)}|w_1^{(1)}) = 1$ and $P(o_1^{(1)}|w_1^{(2)}) = 0$). Finally, the value node $V$, with $O_q$ as sole parent, is associated to a utility function $U(O_q)$

---
[1] Any binary tree decomposition can be transformed to meet this requirement by repeatedly selecting a node $i$ associated to a superset of the parents of a value variable $V_i$ not meeting the requirement, and then adding two nodes $j$ and $k$ such that the children of $i$ become children of $j$, and $k$ is a child of $i$; the node $j$ is associated to the set of variables associated to $i$, while $k$ is associated to $\mathtt{Pa}(V_i)$. Note that the treewidth is unaltered by these operations.

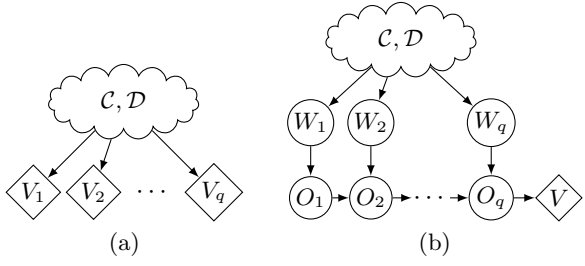

Figure 2: (a) Influence diagram with multiple value variables. (b) Its equivalent influence diagram containing a single variable as used in the proof of Theorem 1.

such that $U(o_q^{(1)}) = q\overline{U}$ and $U(o_q^{(2)}) = q\underline{U}$.

To show that the treewidth of $\mathcal{I}'$ is not greater than the treewidth of $\mathcal{I}$ by more than three, we build a tree decomposition $\mathcal{T}'$ for $\mathcal{I}'$, based on the tree decomposition $\mathcal{T}$, as follows. First, we take each node $\ell_i$ of $\mathcal{T}$ and include $W_i$, $O_i$ and $O_{i-1}$ in $\mathcal{X}_i$ ($O_{i-1}$ is included for $i > 1$), which is enough to cover their families and to satisfy the family preserving property. Note that at this stage $\mathcal{T}'$ does not satisfy the running intersection property, because $O_{i-1}$ appears in both $\ell_{i-1}$ and $\ell_i$ but not necessarily in every (set associated to a) node in between. Finally, we walk around the tree $\mathcal{T}'$ in a Euler tour tree traversal where each edge is visited exactly twice, and we include $O_{i-1}$ in each node of $\mathcal{T}'$ that appears between $\ell_{i-1}$ and $\ell_i$ during the walk.[2] By doing so we guarantee that the running intersection property is also satisfied. Since the Euler tour tree traversal visits each leaf once and each internal node at most three times, the procedure inserts three new variables in the sets associated to $\ell_1, \ldots, \ell_q$ and at most three variables $O_i$ in the sets associated to non-leaf nodes, and therefore does not increase the treewidth of the decomposition by more than three.

It remains to show that the diagrams are equivalent with respect to the expected utility of strategies. Let $\mathcal{S}$ be a strategy for $\mathcal{I}$. Then $\mathcal{S}$ is also a strategy for $\mathcal{I}'$ (because the two diagrams share the same decision variables and graph over $\mathcal{C}, \mathcal{D}$). First, we need to show that for $i = 1, \ldots, q$ it follows that

$$P(o_i^{(1)}|\mathcal{C}, \mathcal{D}) = \frac{1}{i}\left(\sum_{j=1}^{i} P(w_j^{(1)}|\mathcal{C}, \mathcal{D})\right),$$

which we do by induction in $i$. The basis ($i = 1$) follows trivially, because $P(o_1^{(1)}|w_1^{(1)}) = 1$ by definition, so according to the graph structure we have that $P(o_1^{(1)}|\mathcal{C}, \mathcal{D}) = P(w_1^{(1)}|\mathcal{C}, \mathcal{D})$. Now assume by hypothesis of the induction that the above result is valid for every $1 \le i \le k < q$. Then $P(o_{k+1}^{(1)}|\mathcal{C}, \mathcal{D})$

$$= P(o_{k+1}^{(1)}|o_k^{(1)}, w_{k+1}^{(1)})P(o_k^{(1)}|\mathcal{C},\mathcal{D})P(w_{k+1}^{(1)}|\mathcal{C},\mathcal{D})$$
$$+ P(o_{k+1}^{(1)}|o_k^{(1)}, w_{k+1}^{(2)})P(o_k^{(1)}|\mathcal{C},\mathcal{D})P(w_{k+1}^{(2)}|\mathcal{C},\mathcal{D})$$
$$+ P(o_{k+1}^{(1)}|o_k^{(2)}, w_{k+1}^{(1)})P(o_k^{(2)}|\mathcal{C},\mathcal{D})P(w_{k+1}^{(1)}|\mathcal{C},\mathcal{D})$$
$$+ P(o_{k+1}^{(1)}|o_k^{(2)}, w_{k+1}^{(2)})P(o_k^{(2)}|\mathcal{C},\mathcal{D})P(w_{k+1}^{(2)}|\mathcal{C},\mathcal{D})$$
$$= \left(\frac{1}{k+1} + \frac{k}{k+1}\right)P(o_k^{(1)}|\mathcal{C},\mathcal{D})P(w_{k+1}^{(1)}|\mathcal{C},\mathcal{D}) +$$
$$\left(\frac{k}{k+1}\right)P(o_k^{(1)}|\mathcal{C},\mathcal{D})P(w_{k+1}^{(2)}|\mathcal{C},\mathcal{D}) +$$
$$\left(\frac{1}{k+1}\right)P(o_k^{(2)}|\mathcal{C},\mathcal{D})P(w_{k+1}^{(1)}|\mathcal{C},\mathcal{D})$$
$$= \left(\frac{k}{k+1}\right)P(o_k^{(1)}|\mathcal{C},\mathcal{D}) + \frac{1}{k+1}P(w_{k+1}^{(1)}|\mathcal{C},\mathcal{D})$$
$$= \frac{k}{(k+1)}\left(\frac{1}{k}\sum_{j=1}^{k} P(w_j^{(1)}|\mathcal{C},\mathcal{D})\right) + \frac{1}{k+1}P(w_{k+1}^{(1)}|\mathcal{C},\mathcal{D})$$
$$= \frac{1}{k+1}\left(\sum_{j=1}^{k+1} P(w_j^{(1)}|\mathcal{C},\mathcal{D})\right),$$

which shows that the result holds also for $i = k + 1$.

We can now show that for any strategy $\mathcal{S}$ the associated expected utility $\mathrm{E}'_{\mathcal{S}}$ in $\mathcal{I}'$ equals the associated expected utility $\mathrm{E}_{\mathcal{S}}$ in $\mathcal{I}$. Let $\mathcal{C}'' = \mathcal{C}' \setminus \{O_q\}$. By definition, $\mathrm{E}'_{\mathcal{S}} = \sum_{\mathcal{C}',\mathcal{D}} P_{\mathcal{S}}(\mathcal{C}',\mathcal{D})U(\mathcal{C}',\mathcal{D}) = \sum_{\mathcal{C}',\mathcal{D}} P_{\mathcal{S}}(\mathcal{C}',\mathcal{D})U(O_q)$, which is equal to

$$\sum_{\mathcal{C}'',\mathcal{D}} P_{\mathcal{S}}(\mathcal{C}'',\mathcal{D})\left(q\overline{U}P(o_q^{(1)}|\mathcal{C}'',\mathcal{D}) + q\underline{U}P(o_q^{(2)}|\mathcal{C}'',\mathcal{D})\right)$$
$$= q\underline{U} + \sum_{\mathcal{C}'',\mathcal{D}} P_{\mathcal{S}}(\mathcal{C}'',\mathcal{D})q(\overline{U} - \underline{U})P(o_q^{(1)}|\mathcal{C}'',\mathcal{D})$$
$$= q\underline{U} + q(\overline{U} - \underline{U})\sum_{\mathcal{C},\mathcal{D}} P_{\mathcal{S}}(\mathcal{C},\mathcal{D})P(o_q^{(1)}|\mathcal{C},\mathcal{D})$$
$$= q\underline{U} + q(\overline{U} - \underline{U})\sum_{\mathcal{C},\mathcal{D}} P_{\mathcal{S}}(\mathcal{C},\mathcal{D})\frac{1}{q}\sum_{i=1}^{q} P(w_i^{(1)}|\mathcal{C},\mathcal{D})$$
$$= q\underline{U} + q\frac{\overline{U} - \underline{U}}{q}\sum_{\mathcal{C},\mathcal{D}} P_{\mathcal{S}}(\mathcal{C},\mathcal{D})\sum_{i=1}^{q} P(w_i^{(1)}|\mathtt{Pa}(V_i))$$
$$= q\underline{U} + (\overline{U} - \underline{U})\sum_{\mathcal{C},\mathcal{D}} P_{\mathcal{S}}(\mathcal{C},\mathcal{D})\sum_{j=1}^{q} \frac{U(\mathtt{Pa}(V_i)) - \underline{U}}{\overline{U} - \underline{U}}$$
$$= q\underline{U} + \frac{\overline{U} - \underline{U}}{\overline{U} - \underline{U}}\sum_{\mathcal{C},\mathcal{D}} P_{\mathcal{S}}(\mathcal{C},\mathcal{D})\sum_{i=1}^{q} [U(\mathtt{Pa}(V_i)) - \underline{U}]$$
$$= q\underline{U} - q\underline{U} + \sum_{\mathcal{C},\mathcal{D}} P_{\mathcal{S}}(\mathcal{C},\mathcal{D})\sum_{i=1}^{q} U(\mathtt{Pa}(V_i)),$$

---

[2] An Euler tour tree traversal of $\mathcal{T}$ rooted at $r$ is a list of $2m-1$ symbols produced by calling $ET(r)$, where $ET(i)$ is a recursive function that takes a node $i$ with left children $j$ and right children $k$ (if they exist) and prints out $i$, calls $ET(j)$ (if $j$ exists), prints out $i$ again, calls $ET(k)$ (if $k$ exists), and then prints out $i$ once more.

which is equal to $\sum_{\mathcal{C},\mathcal{D}} P_\mathcal{S}(\mathcal{C},\mathcal{D})U(\mathcal{C},\mathcal{D}) = \mathrm{E}_\mathcal{S}$. □

Cooper [3] showed that in any influence diagram containing a single value variable the (single) utility function can be re-normalized to take values on the interval $[0, 1]$ without affecting the optimality of the strategies. Together with the above result, this allows us, without loss of generality, to focus exclusively on influence diagrams containing a single value variable taking values on the interval $[0, 1]$.

Under the usual assumptions of complexity theory, when a problem is NP-hard to solve the best available options are (i) trying to devise an algorithm that runs efficiently on many instances but has exponential worst-case complexity, or (ii) trying to develop an approximation algorithm that for all instances provides in polynomial time a solution that is provably within a certain range of the optimal solution. We have investigated option (i) in a recent work [11, 12], and showed empirically that a sophisticate variable elimination algorithm can, in spite of its exponential worst-case complexity, solve a large number of problems in feasible time. In the remainder, we study the theoretical feasibility of option (ii). Given $0 < \epsilon < 1$, we say that a procedure is an $\epsilon$-approximation algorithm for the strategy selection problem if for any influence diagram it finds a strategy $\mathcal{S}$ such that $\mathrm{MEU} \leq (1+\epsilon)\mathrm{E}_\mathcal{S}$. The algorithm is said to be a *fully polynomial-time approximation scheme* if it runs in time polynomial in the input and in $1/\epsilon$ for any given $\epsilon$. The following result implies that without assuming a bounded number of states per variable approximating the strategy selection problem in polynomial time by virtually any reasonable factor is NP-hard, and option (ii) is most likely unfeasible.

**Theorem 2** ([12]). *Given a singly connected influence diagram of bounded treewidth, for any $1 < 1 + \epsilon < 2^\theta$, there is no polynomial-time $\epsilon$-approximation algorithm unless P=NP, where $\theta$ is the number of numerical parameters (i.e., probabilities and rewards) needed to specify the problem.*

The result is obtained by a reduction from the SAT problem, similarly to the proof of inapproximability of MAP in Bayesian networks given by Park and Darwiche [16]. According to the above result, algorithms like SPU [10] or mini-bucket elimination [5], which find a strategy in polynomial time, cannot guarantee that the worst-case ratio between the expected utility of that strategy and the maximum expected utility is greater than $2^{-\theta}$, even if the input is constrained to influence diagrams of bounded treewidth. Since the factor $2^{-\theta}$ quickly approaches zero as the size of the problem increases, efficient algorithms eventually produce very poor solutions. As we show in the rest of the paper, this is no longer true if we assume that both the treewidth of the diagram and the cardinality of the variables are bounded. In this case, we show constructively the existence of a fully polynomial-time approximation scheme.

## 3  FINDING PROVABLY GOOD STRATEGIES

Our first step in showing the existence of a fully polynomial-time approximation scheme for the strategy selection problem is to devise an algorithm that obtains provably good strategies, that is, an $\epsilon$-approximation scheme which returns, for any $\epsilon > 0$, a strategy $\mathcal{S}$ satisfying $\mathrm{MEU} \leq (1+\epsilon)\mathrm{E}_\mathcal{S}$. The algorithm (scheme) we devise is a generalization of the MPU algorithm [11, 12] to the case of (provably good) approximate inference in influence diagrams, and it consists in propagating sets of probability potentials over a tree decomposition. The idea behind the propagation of sets of potentials is that the expected utility of each strategy can be computed by propagating a (single) potential over the same tree decomposition, and so the propagated sets account for the simultaneous evaluation of many different strategies. To be efficient, some of these computations are halted early on, so that not every strategy has its expected value computed. The difficulty is in guaranteeing that within the evaluated strategies (i.e., those whose expected values were actually computed) there is at least one which achieves the desired approximation factor $1 + \epsilon$.

Let $\alpha$ be a real value greater than one, and $\mathcal{K}$ be a set of nonnegative functions (called potentials) over a set of variables $\mathcal{X}$. We say that $\mathcal{K}' \subseteq \mathcal{K}$ is a $\alpha$-covering of $\mathcal{K}$ if for every potential $P(\mathcal{X})$ in $\mathcal{K}$ there is $Q(\mathcal{X})$ in $\mathcal{K}'$ such that $P(\mathcal{X}) \leq \alpha Q(\mathcal{X})$. We also define the potential $1(\mathcal{X})$ which returns one for every assignment to the variables in $\mathcal{X}$.

For any decision variable $D$, let $\mathcal{P}_D$ denote the set of all pure policies for $D$. Hence, $\mathcal{P}_D$ is a set containing $|D|^\gamma$ distributions $P(D|\mathtt{Pa}(D))$, where $|D|$ is the number of values $D$ can assume and $\gamma = \prod_{X \in \mathtt{Pa}(D)} |X|$ is the number of assignments to its parents (i.e., the number of different combinations of values the parent variables can jointly assume).

The procedure in Algorithm 1 takes an influence diagram (which we assume contains a single value variable taking values on $[0, 1]$), a tree decomposition of the diagram, and a nonnegative value $\epsilon$, and computes an expected utility $E_\mathcal{S}$ induced by some strategy $\mathcal{S}$ such that $\mathrm{MEU} \leq (1 + \epsilon)E_\mathcal{S}$, that is, it is an $\epsilon$-approximation for computing the maximum expected utility. To keep things simple for now, let us assume

**Algorithm 1** Finding Provably Good Strategies

**Require:** A positive number $\epsilon$, an influence diagram $\mathcal{I}$ and a tree decomposition $\mathcal{T}$ with $m$ nodes
**Ensure:** The value E satisfies MEU $\leq (1+\epsilon)$ E
1: // INITIALIZATION
2: let $\alpha = 1 + \epsilon/(2m)$
3: let $\mathcal{K}_1, \ldots, \mathcal{K}_m$ be singletons containing the potentials $1(\mathcal{X}_1), \ldots, 1(\mathcal{X}_m)$, respectively
4: **for each** chance variable $C$ **do**
5:     find a node $i$ in $\mathcal{T}$ such that $\texttt{Fa}(C) \subseteq \mathcal{X}_i$
6:     set $\mathcal{K}_i \leftarrow \textsc{Combine}(\mathcal{K}_i, \{P(C|\texttt{Pa}(C))\})$
7: **end for**
8: **for each** decision variable $D$ **do**
9:     find a node $i$ in $\mathcal{T}$ such that $\texttt{Fa}(D) \subseteq \mathcal{X}_i$
10:    set $\mathcal{K}_i \leftarrow \textsc{Combine}(\mathcal{K}_i, \mathcal{P}_D)$
11: **end for**
12: find a node $i$ in $\mathcal{T}$ such that $\texttt{Pa}(V) \subseteq \mathcal{X}_i$
13: set $\mathcal{K}_i \leftarrow \textsc{Combine}(\mathcal{K}_i, \{U(\texttt{Pa}(V))\})$
14: // PROPAGATION
15: select a node $r$ as the root of $\mathcal{T}$
16: label all nodes as inactive
17: **while** there is an inactive node $i$ **do**
18:    select an inactive node $i$ whose children are all active
19:    set $\mathcal{A}_i \leftarrow \textsc{Combine}(\mathcal{K}_i, \mathcal{C}_j : j \in \texttt{Ch}(i))$
20:    set $\mathcal{B}_i \leftarrow \textsc{SumOut}(\mathcal{A}_i, \mathcal{X}_i \setminus \mathcal{X}_{\texttt{Pa}(i)})$
21:    set $\mathcal{C}_i \leftarrow \textsc{Covering}(\mathcal{B}_i, \alpha)$
22:    label $i$ as active
23: **end while**
24: let $\text{E} = \max\{\mu : \mu \in \mathcal{C}_r\}$

---

this is our goal. We will see later on how to modify the algorithm to also provide the strategy $\mathcal{S}$, not only its expected value, and thus devise an $\epsilon$-approximation algorithm for the strategy selection problem.

The algorithm is similar in spirit to the computation of marginal queries in Bayesian networks by junction trees [8], but differs radically in that it stores and propagates sets of probability potentials, instead of storing and propagating single probability potentials. Like junction-tree algorithms for Bayesian networks, the algorithm contains an initialization step, where the sets of probability potentials are assigned to nodes of the tree decomposition, and a propagation step, where messages are sent from the leaves towards the root node. The propagation finishes when the root receives a message from every child, and obtains a set of values $\mathcal{C}_r$ (since $\texttt{Pa}(r) = \emptyset$, the set $\mathcal{B}_r$ marginalizes out all variables from each potential $P(\mathcal{X}_r)$ in $\mathcal{B}_r$, hence producing real numbers).

The operations $\textsc{Combine}$ and $\textsc{SumOut}$ are analogous to the multiplication and marginalization of probability potentials but operate over sets, that is, the former returns the set of all potentials obtained by pairwise multiplication of the potentials in the sets given as arguments, whereas the latter returns the set obtained by summing out the variables in the second argument from every potential in the first argument. More formally, given a list of sets $\mathcal{K}_1, \ldots, \mathcal{K}_n$ containing potentials over the sets of variables $\mathcal{Y}_1, \ldots, \mathcal{Y}_n$, respectively, we define the $\textsc{Combine}$ operation as $\textsc{Combine}(\mathcal{K}_1, \ldots, \mathcal{K}_n) = \{P(\mathcal{Y}_1) \cdots P(\mathcal{Y}_n) : P(\mathcal{Y}_i) \in \mathcal{K}_i, i = 1, \ldots, n\}$; given a set $\mathcal{K}$ of potentials over the set of variables $\mathcal{Y}$ and a subset $\mathcal{Z} \subseteq \mathcal{Y}$, the $\textsc{SumOut}$ operation is given by $\textsc{SumOut}(\mathcal{K}, \mathcal{Z}) = \{\sum_{\mathcal{Z}} P(\mathcal{Y}) : P(\mathcal{Y}) \in \mathcal{K}\}$. Finally, the $\textsc{Covering}$ operation returns an $\alpha$-covering for the argument, where $\alpha = 1 + \epsilon/(2m)$ is set according to the approximation factor $\epsilon$ and the number of nodes in the tree $m$. For the moment, let us ignore how $\alpha$-coverings are actually obtained. We shall get back to this point later on.

The algorithm begins by assigning each probability distribution associated to a chance variable to a node of the tree decomposition. If two or more such functions are assigned to the same node $i$, the result is a singleton $\mathcal{K}_i$ containing their multiplication. The algorithm then assigns policies to nodes in much a similar way, except that the sets associated to nodes which have been chosen in the second loop are no longer singletons. Finally, the utility function is associated to a node of the tree. For example, if the tree decomposition contained only a single node with all decision and chance variables associated to it, the result of the initialization step would be a single set $\mathcal{K}_1$ containing all joint probability distributions $P_\mathcal{S}(\mathcal{C}, \mathcal{D})$ induced by strategies $\mathcal{S}$. On the other hand, if for each (decision, chance and value) variable $X$ there were exactly one node $i$ in the tree such that $\texttt{Fa}(X) \subseteq \mathcal{X}_i$, then after the initialization step the set $\mathcal{K}_i$, for $i = 1, \ldots, m$, would be equal to the singleton $\{P(X|\texttt{Pa}(X))\}$, if the node $i$ were associated to the family of a chance variable $X$, to the singleton $\{U(\texttt{Pa}(X))\}$, if $i$ were the node associated to the parents of the value variable $V$, or to the set of policies $\mathcal{P}_X$, if $i$ was associated to the family of a decision variable $X$.

The propagation step starts by rooting the tree decomposition $\mathcal{T}$ at an arbitrary node $r$. This allows us to organize the neighbors of a node in the tree as the parent (i.e., the one closer to the root) and the children (the ones farther from the root), and defines a direction for the propagation of messages, which consists in selecting a node $i$ satisfying the condition (which initially only leaves do) and computing the corresponding sets of potentials $\mathcal{A}_i$, $\mathcal{B}_i$ and $\mathcal{C}_i$. The first two are set analogous to the potentials produced during a junction-tree propagation in a Bayesian network. The message set $\mathcal{C}_i$ is obtained by removing elements from

$\mathcal{B}_i$ in a way that satisfies the $\alpha$-covering condition (we will get back to this). Once all nodes have been processed (including the root), the algorithm produces a solution E by seeking the highest value in the set $\mathcal{C}_r$. To show that this value is indeed the expected utility of a feasible strategy $\mathcal{S}$ such that $\mathrm{MEU} \leq (1+\epsilon)\,\mathrm{E}_{\mathcal{S}}$, we need to introduce some additional notation.

For any node $i$ of the tree decomposition, let $\mathcal{T}[i]$ be the subtree rooted at $i$, that is, the graph obtained by removing all nodes which are not descendants of $i$ or $i$ itself. Let also $\mathcal{X} = \mathcal{X}_1 \cup \cdots \cup \mathcal{X}_m = \mathcal{C} \cup \mathcal{D}$ denote all variables in the tree. We denote by $\mathcal{X}[i]$ the set of all variables associated to the nodes in $\mathcal{T}[i]$, that is, $\mathcal{X}[i] = \bigcup_{j \in \mathcal{T}[i]} \mathcal{X}_j$ (hence $\mathcal{X}_i \subseteq \mathcal{X}[i]$). We define a function $\sigma : \mathcal{X} \to \{1, \ldots, m\}$ that returns for each variable $X$ the node $i$ to which its associated (probability, policy or utility) potential was assigned in the initialization step. Given a node $i$, the function $U(\mathtt{Pa}(V); i)$ returns the utility function $U(\mathtt{Pa}(V))$ if $\sigma(V) = i$ and one otherwise. Finally, let $\mathcal{Y}_i = \mathcal{X}_i \cap \mathcal{X}_{\mathtt{Pa}(i)}$ be the separator set of $i$ and $\mathtt{Pa}(i)$.

**Lemma 3.** *The value E computed by the procedure in Algorithm 1 is the expected utility $\mathrm{E}_{\mathcal{S}}$ associated to some valid strategy $\mathcal{S}$.*

*Proof.* First we show by induction in the nodes from the leaves toward the root that for any node $i$ any potential $P(\mathcal{Y}_i)$ in $\mathcal{B}_i$ or $\mathcal{C}_i$ satisfies

$$P(\mathcal{Y}_i) = \sum_{\mathcal{X}[i] \setminus \mathcal{X}_{\mathtt{Pa}(i)}} \prod_{C:\sigma(C) \in \mathcal{T}[i]} P(C|\mathtt{Pa}(C)) \prod_{D:\sigma(D) \in \mathcal{T}[i]} P(D|\mathtt{Pa}(D)) \prod_{j \in \mathcal{T}[i]} U(\mathtt{Pa}(V); j)$$

for some partial strategy $\{P(D|\mathtt{Pa}(D)) : D \in \mathcal{D}, \sigma(D) \in \mathcal{T}[i]\}$. Consider a leaf node $i$. Then the induction hypothesis is trivially satisfied by applying the definitions of the COMBINE and SUMOUT operations, and noting that $\mathcal{T}[i]$ contains only the node $i$. Now consider an internal node $i$ and assume that for every child $j$ of $i$ the induction hypothesis holds. Then for any $P(\mathcal{Y}_i)$ in $\mathcal{B}_i$ or in $\mathcal{C}_i$ it follows that $P(\mathcal{Y}_i)$

$$= \sum_{\mathcal{X}_i \setminus \mathcal{X}_{\mathtt{Pa}(i)}} \prod_{C:\sigma(C)=i} P(C|\mathtt{Pa}(C)) \prod_{D:\sigma(D)=i} P(D|\mathtt{Pa}(D))$$
$$U(\mathtt{Pa}(V); i) \prod_{j \in \mathtt{Ch}(i)} P(\mathcal{Y}_j)$$
$$= \sum_{\mathcal{X}[i] \setminus \mathcal{X}_{\mathtt{Pa}(i)}} \prod_{C:\sigma(C) \in \mathcal{T}[i]} P(C|\mathtt{Pa}(C))$$
$$\prod_{D:\sigma(D) \in \mathcal{T}[i]} P(D|\mathtt{Pa}(D)) \prod_{j \in \mathcal{T}[i]} U(\mathtt{Pa}(V); j),$$

which satisfies the induction hypothesis. The result of the lemma is thus obtained by applying the induction result to the root node $r$. For any $\mu \in \mathcal{C}_r$ we have that

$$\mu = \sum_{\mathcal{X}} \prod_{C} P(C|\mathtt{Pa}(C)) \prod_{D} P(D|\mathtt{Pa}(D)) U(\mathtt{Pa}(V))$$

for some strategy $S = \{P(D|\mathtt{Pa}(D))\}$. $\square$

**Theorem 4.** *The value E satisfies $\mathrm{MEU} \leq (1+\epsilon)\,\mathrm{E}$.*

*Proof.* Let $S^* = \{P^*(D|\mathtt{Pa}(D)) : D \in \mathcal{D}\}$ denote an optimal strategy. We first show by induction from the leaves toward the root that for any node $i$ there is a $P(\mathcal{Y}_i) \in \mathcal{C}_i$ such that $P^*(\mathcal{Y}_i) \leq \alpha^{s_i} P(\mathcal{Y}_i)$, where

$$P^*(\mathcal{Y}_i) = \sum_{\mathcal{X}[i] \setminus \mathcal{X}_{\mathtt{Pa}(i)}} \prod_{C:\sigma(C) \in \mathcal{T}[i]} P(C|\mathtt{Pa}(C))$$
$$\prod_{D:\sigma(D) \in \mathcal{T}[i]} P^*(D|\mathtt{Pa}(D)) \prod_{j \in \mathcal{T}[i]} U(\mathtt{Pa}(V); j),$$

and $s_i$ is the number of nodes in $\mathcal{T}[i]$. Consider a leaf node $i$. Then the induction hypothesis holds since $P^*(\mathcal{Y}_i) \in \mathcal{B}_i$ by design and by definition of $\alpha$-covering there is $P(\mathcal{Y}_i) \in \mathcal{C}_i$ such that $P^*(\mathcal{Y}_i) \leq \alpha P(\mathcal{Y}_i)$. Assume the induction holds for all children $j$ of a node $i$. By design, there is $P(\mathcal{Y}_i) \in \mathcal{B}_i$ such that

$$P(\mathcal{Y}_i) = \sum_{\mathcal{X}_i \setminus \mathcal{X}_{\mathtt{Pa}(i)}} \prod_{C:\sigma(C)=i} P(C|\mathtt{Pa}(C))$$
$$\prod_{D:\sigma(D)=i} P^*(D|\mathtt{Pa}(D)) U(\mathtt{Pa}(V); i) \prod_{j \in \mathtt{Ch}(i)} P(\mathcal{Y}_j),$$

and by using the inductive hypothesis it follows that $P^*(\mathcal{Y}_i) \leq \alpha^{\sum_{j \in \mathtt{Ch}(i)} s_j} P(\mathcal{Y}_i)$. And since $\mathcal{C}_i$ is an $\alpha$-covering of $\mathcal{B}_i$, there is $Q(\mathcal{Y}_i) \in \mathcal{C}_i$ such that $P(\mathcal{Y}_i) \leq \alpha Q(\mathcal{Y}_i)$, and thus $P^*(\mathcal{Y}_i) \leq \alpha^{1+\sum_{j \in \mathtt{Ch}(i)} s_j} Q(\mathcal{Y}_i) = \alpha^{s_i} Q(\mathcal{Y}_i)$. Since by Lemma 3, every $\mu \in \mathcal{C}_r$ corresponds to the expected utility of some strategy $\mathcal{S}$, it follows from the induction result on $r$ that there is $\mathrm{E}_{\mathcal{S}} \in \mathcal{C}_r$ such that $\mathrm{MEU} \leq \alpha^m \mathrm{E}_{\mathcal{S}}$, and since E maximizes over $\mu \in \mathcal{C}_r$, we have that $\mathrm{MEU} \leq \alpha^m \mathrm{E} = (1 + \epsilon/(2m))^m \mathrm{E}$. Finally, it follows from the inequality $(1 + 2x) \geq (1 + x/k)^k$, valid for every positive real $x \leq 1$ and positive integer $k$, that $\mathrm{MEU} \leq (1+\epsilon)\,\mathrm{E}$. $\square$

Recall from Section 2 that for any influence diagram of bounded treewidth we can obtain a binary tree decomposition of minimum treewidth in linear time. Assume, without loss of generality, that such a minimum treewidth binary tree decomposition is given as input to the approximation algorithm, and let $\omega$ be the treewidth of the influence diagram given as input and $\kappa$ be the maximum number of values a decision or chance variable can assume. The complexity of the algorithm is bounded by the complexity of computing a potential $P(\mathcal{Y}_i)$ in a set $\mathcal{B}_i$ times the cardinality of the largest set $\mathcal{B}_i$ plus the complexity of the initialization step and the complexity of the COVERING operation. Computing a $P(\mathcal{Y}_i)$ requires $2\prod_{X\in\mathcal{X}_i}|X|$ multiplications and $\prod_{X\in\mathcal{X}_i\setminus X_{\text{Pa}(i)}}|X|$ additions, and hence takes $O(\kappa^\omega)$ time, which is polynomial in the cardinality of the (chance and decision) variables (since $\omega$ is assumed to be bounded by a constant). The complexity of the initialization step is bounded by the complexity of combining the sets of policies $\mathcal{P}_D$ with the sets $\mathcal{K}_i$ for each decision variable. Each decision variable has at most $\kappa^{\kappa^\omega}$ policies, which makes this step exponential in $\kappa$. Although it is possible to transform the diagram so that the complexity of the initialization step becomes polynomially bounded in $\kappa$ [12, Prop. 7], we refrain from doing so here because it would make the algorithm more complicated, and it would not change the worst-case running time, which we know from Theorem 2 that cannot be polynomial in $\kappa$.

Regarding the cardinality of the sets $\mathcal{C}_i$ and the complexity of the COVERING operations, in principle, we would like to be able to implement COVERING in way that it takes polynomial time in its argument and provides an $\alpha$-covering that is as small as possible. The former requirement is easily met by sequentially inspecting an element $P(\mathcal{Y}_i)$ in $\mathcal{B}_i$ and inserting it into $\mathcal{C}_i$ only if there is no other $Q(\mathcal{Y}_i)$ in $\mathcal{B}_i$ such that $P(\mathcal{Y}_i) \leq \alpha Q(\mathcal{Y}_i)$. This procedure takes time polynomial in the cardinality of $\mathcal{B}_i$ and in $|\mathcal{Y}_i|$, but does not guarantee that the size of the obtained set is bounded, and in the worst case we might simply output $\mathcal{C}_i = \mathcal{B}_i$ at every node $i$, which would cause $\mathcal{C}_r$ to contain as many as $\prod_{i=1}^m |\mathcal{K}|_i$ or $O(\kappa^{\kappa^{m\omega}})$ elements. As we show in the next section, there is a better way of finding $\alpha$-coverings (in polynomial time) that guarantees that the cardinality of the sets $\mathcal{C}_i$ remains bounded in the (size of the influence diagram given as) input (measured in bits and using a reasonable encoding) if the cardinality of any variable is bounded.

Before introducing the fully polynomial-time approximation scheme, there is a minor detail we have postponed in the discussion, which is how to modify the algorithm to provide not only the expected value but also a strategy that obtains that value. This can be easily implemented by associating to any potential generated during the algorithm the policies that were used either directly or indirectly to produce it. More specifically, let $\delta$ be a dictionary, which is initialized with $\delta[1(\mathcal{X}_i)] \leftarrow \{\}$ for $i = 1, \ldots, m$, $\delta[P(C|\text{Pa}(C))] \leftarrow \{\}$ for every $C$ in $\mathcal{C}$, $\delta[U(\text{Pa}(V))] \leftarrow \{\}$, and $\delta[P(D|\text{Pa}(D))] \leftarrow \{P(D|\text{Pa}(D))\}$ for every $D$ in $\mathcal{D}$ and $P(D|\text{Pa}(D))$ in $\mathcal{P}_D$. We redefine the operations COMBINE and SUMOUT to update $\delta$ as the computations are done as follows. Let $\mathcal{K}$ be the output of COMBINE($\mathcal{K}_1, \ldots, \mathcal{K}_n$). Then for every $P(\mathcal{Y}_1)\cdots P(\mathcal{Y}_n) \in \mathcal{K}$, where $P(\mathcal{Y}_1) \in \mathcal{K}_1, \ldots, P(\mathcal{Y}_n) \in \mathcal{K}_n$, we assign $\delta[P(\mathcal{Y}_1)\cdots P(\mathcal{Y}_n)] \leftarrow \delta[P(\mathcal{Y}_1)] \cup \cdots \cup \delta[P(\mathcal{Y}_n)]$. Likewise, let $\mathcal{K}$ be the output of SUMOUT($\mathcal{K}', \mathcal{Z}$). Then $\delta[\sum_{\mathcal{Z}} P(\mathcal{Y})] \leftarrow \delta[P(\mathcal{Y})]$ for every $\sum_{\mathcal{Z}} P(\mathcal{Y})$ in $\mathcal{K}$, where $P(\mathcal{Y}) \in \mathcal{K}'$. The COVERING operation does not need modification since it returns a subset of its argument. Finally, we obtain a strategy $\mathcal{S}$ such that $E_{\mathcal{S}} = E$ from $\delta[E]$. Note that these additional operations do not change the asymptotical running time complexity, and can be efficiently implemented using pointers to the original functions, incurring a very small increase in the space complexity.

# 4 A FULLY POLYNOMIAL-TIME APPROXIMATION SCHEME

When the maximum cardinality of a variable $\kappa$ is assumed bounded, the complexity of computing each potential $P(\mathcal{Y}_i)$ in the procedure in Algorithm 1 as well as the cardinalities of the sets $\mathcal{K}_i$ are bounded by a constant. Hence, the only difficulty one needs to overcome in order to devise a fully polynomial-time approximation scheme out of that procedure is to guarantee that the operation COVERING returns sets $\mathcal{C}_i$ whose cardinality is polynomially bounded by the input size, where the latter is defined as the number of bits needed to encode all numerical parameters (i.e., probabilities and utilities) as well as the variables and the underlying graph of the diagram. For definiteness, we assume the numerical parameters are specified as rational numbers in a reasonable way. The procedure in Algorithm 2 partitions the space $\mathcal{Y}$ over which the potentials in the input set $\mathcal{K}$ are specified in hyper-rectangles such that the ratio of any two potentials falling in the same rectangle is at most $\alpha$. Thus, we can provide an $\alpha$-covering of $\mathcal{K}$ by letting $\mathcal{K}'$ be a set obtained by selecting exactly one element from each non-empty rectangle.[3] Remarkably, we show that the number of elements in $\mathcal{K}'$ is a polynomial function of the size of the influence diagram (in bits).

The next result, which is inspired by a similar result by Papadimitriou and Yannakakis [15], relates the cardinality of the output of the COVERING procedure in Algorithm 2 with the size in bits of the input set $\mathcal{K}$.

**Lemma 5.** *Let $\mathcal{K}$ be a set of potentials $P(\mathcal{Y})$ whose range is contained in $[0, 1]$. Then the set $\mathcal{K}'$ obtained by the procedure in Algorithm 2 is an $\alpha$-covering of $\mathcal{K}$ with at most $(1 - \lfloor\log_\alpha t\rfloor)^\eta$ elements, where $t$ is the smallest (strictly) positive number in the range of a potential in $\mathcal{K}$ and $\eta = \prod_{X\in\mathcal{Y}}|X|$ is the number of assignments to $\mathcal{Y}$.*

---

[3]The notation $\lfloor\log_\alpha P(\mathcal{Y})\rfloor$ denotes a function $F(\mathcal{Y})$ such that $F(\boldsymbol{y}) = \lfloor\log_\alpha P(\boldsymbol{y})\rfloor$ if $P(\boldsymbol{y}) \neq 0$ and $F(\boldsymbol{y}) = 0$ otherwise, where $\boldsymbol{y}$ is an assignment to $\mathcal{Y}$.

**Algorithm 2** Finding A Small $\alpha$-Covering

**Require:** A set $\mathcal{K}$ of potentials over a set of variables $\mathcal{Y}$, a value $\alpha > 1$
**Ensure:** $\mathcal{K}'$ is an $\alpha$-covering of $\mathcal{K}$
1: let $\mathcal{K}'$ and $\mathcal{L}$ initially be empty sets
2: **for each** $P(\mathcal{Y})$ in $\mathcal{K}$ **do**
3:    **if** $\lfloor \log_\alpha P(\mathcal{Y}) \rfloor$ is not in $\mathcal{L}$ **then**
4:       insert $\lfloor \log_\alpha P(\mathcal{Y}) \rfloor$ into $\mathcal{L}$
5:       insert $P(\mathcal{Y})$ into $\mathcal{K}'$
6:    **end if**
7: **end for**

*Proof.* To see that $\mathcal{K}'$ is an $\alpha$-covering of $\mathcal{K}$, note that by design if a potential $P(\mathcal{Y}) \in \mathcal{K}$ is not in $\mathcal{K}'$ then the latter contains a potential $Q(\mathcal{Y}) \in \mathcal{K}$ such that $\lfloor \log_\alpha P(\mathcal{Y}) \rfloor = \lfloor \log_\alpha Q(\mathcal{Y}) \rfloor$, which implies $P(\mathcal{Y}) \leq \alpha Q(\mathcal{Y})$.

Regarding the cardinality of $\mathcal{K}'$, first note that there are $-\lfloor \log_\alpha t \rfloor$ (distinct) integers between $t$ and one (the minus sign is because $t \leq 1$). Now consider a potential $P(\mathcal{Y}) \in \mathcal{K}$. By assumption, the range of $P(\mathcal{Y})$ is contained in $[0, 1]$, and for each assignment $\boldsymbol{y}$ to $\mathcal{Y}$ we have that either $P(\boldsymbol{y}) \geq t$ or $P(\boldsymbol{y}) = 0$. Hence, for each $\boldsymbol{y}$ there are only $(1 - \lfloor \log_\alpha t \rfloor)$ distinct values the number $\lfloor \log_\alpha P(\boldsymbol{y}) \rfloor$ can assume, and therefore only $(1 - \lfloor \log_\alpha t \rfloor)^\eta$ possibilities for $\lfloor \log_\alpha P(\mathcal{Y}) \rfloor$. □

The following result is an immediate consequence of the above result which shows that the cardinality of any set $\mathcal{C}_i$ is polylogarithmic in the smallest positive number being specified by a potential in $\mathcal{B}_i$.

**Corollary 6.** *For $i = 1, \ldots, m$, the set $\mathcal{C}_i$ contains $O([1 - \lfloor \log_\alpha t_i \rfloor]^{\kappa^\omega})$ elements, where $t_i$ is the smallest (strictly) positive number in a potential in $\mathcal{B}_i$.*

We can now state the main result of this paper.

**Theorem 7.** *There is a fully polynomial-time approximation scheme for influence diagrams of bounded treewidth and bounded variable cardinality.*

*Proof.* Assume, without loss of generality, that the influence diagram given as input contains only one value variable taking values in $[0, 1]$. Let $t$ denote the smallest (strictly) positive numerical parameter in the specification (i.e., the smallest nonzero probability or utility specified by the diagram), and let $b$ denote the size of the diagram (in bits). Since the input probabilities and utilities are (by assumption) rational numbers, each positive input number is not smaller than $2^{-b}$ (otherwise we would need more than $b$ bits to encode it). Any potential $P(\mathcal{Y})$ in a set $\mathcal{B}_i$ is obtained by multiplying and marginalizing the functions specified by the diagram, and hence any value $P(\boldsymbol{y})$ in $P(\mathcal{Y})$ is a polynomial in the numerical parameters in the input. Moreover, since each variable in the network is associated to a function over at most $\kappa^{\kappa^\omega}$ numbers, the polynomial has degree at most $O(n\kappa^{\kappa^\omega}) \leq O(n)$, where $n$ is the number of variables. In particular, the smallest positive value $t_i$ in a potential in $\mathcal{B}_i$, for $i = 1, \ldots, m$, is also a polynomial in the numerical parameters of the input of degree $O(n)$, and since these are either zero or some number greater than or equal to $2^{-b}$, it follows that $t_i \geq 2^{-bO(n)}$. Thus, we have from Corollary 6 that the cardinality of any set $\mathcal{C}_i$ is $O([1 - \lfloor \log_\alpha 2^{-bO(n)} \rfloor]^{\kappa^\omega}) \leq O([bn/\ln(\alpha)]^{\kappa^\omega})$. But since $\alpha = 1 + \epsilon/2m$, we have from the inequality $\ln(1 + x) \geq x/(1 + x)$ valid for all $x > 0$ that

$$O\left(\left[\frac{bn}{\ln(\alpha)}\right]^{\kappa^\omega}\right) \leq O\left(\left[bn \frac{1 + \frac{\epsilon}{2n}}{\frac{\epsilon}{2n}}\right]^{\kappa^\omega}\right) \leq O\left(\left[\frac{bn^2}{\epsilon}\right]^{\kappa^\omega}\right).$$

Hence, for $i = 1, \ldots, m$ the number of elements in any $\mathcal{C}_i$ is polynomial in $b$, $n$ and $1/\epsilon$. □

## 4.1 CONCLUSION

Influence diagrams provide a very expressive language to describe decision problems under uncertainty, especially if no-forgetting and regularity are not required. Finding an optimal strategy for such problems is NP-hard even in diagrams of bounded treewidth and very simple structure (e.g., a tree), which makes approximation algorithms an interesting alternative. Here again the problem shows itself to be hard. Without assuming that variables have bounded cardinality, there is no polynomial-time approximation algorithm unless P=NP. As we show here, neither restricting the diagrams to a single value variable makes the problem easier to solve or approximate.

In this paper, we give some hope in light of so many negative results by showing that when the diagram has bounded treewidth and the variables take on a bounded number of values, there is a fully polynomial-time approximation algorithm for the (optimal) strategy selection problem. Although our proof is constructive, the algorithm we provide is not expected to be practical for any reasonably large problem due to the huge constants hidden in the asymptotic analysis. Nevertheless, the existence of such an scheme shall motivate researchers to investigate more efficient approximation algorithms to solve influence diagrams of low treewidth and low variable cardinality.

### Acknowledgements

This work was partially supported by the Swiss National Science Foundation grants no. 200020_134759/1 and 200020_137680/1, and Hasler Foundation grant no. 10030.